\documentclass[runningheads]{llncs}
\usepackage[utf8]{inputenc}
\usepackage{graphicx}
\usepackage{amsmath}
\usepackage{float}
\usepackage{enumitem}
\usepackage{eurosym}
\usepackage{algorithm}
\usepackage{algorithmic}
\usepackage{booktabs}
\usepackage{dblfloatfix}
\usepackage{todonotes}
\usepackage{tabularx}
\usepackage[normalem]{ulem}
\usepackage[capitalise]{cleveref}

\makeatletter

\newcommand{\printfnsymbol}[1]{%
  \textsuperscript{\@fnsymbol{#1}}%
}

\begin{document}
\frontmatter          
\mainmatter              
\title{Utilizing Temporal Information in Deep Convolutional Network for Efficient Soccer Ball Detection and Tracking}
%
\titlerunning{Soccer Ball Detection}  
\author{Anna Kukleva\thanks{Equal Contribution} \and Mohammad Asif Khan\printfnsymbol{1} \and Hafez Farazi \and Sven Behnke}
\authorrunning{A. Kukleva, A. Khan et al.}   

\institute{Universit\"at Bonn, Computer Science Institute VI, Autonomous Intelligent Systems, Endenicher Allee 19a, 53115 Bonn, Germany\\
\email{ \{s6ankukl, s6mokhan\}@uni-bonn.de, \{farazi, behnke\}@ais.uni-bonn.de}}

\maketitle

\begin{abstract}
Soccer ball detection is identified as one of the critical challenges in the RoboCup competition. It requires an efficient vision system capable of handling the task of detection with high precision and recall and providing robust and low inference time. In this work, we present a novel convolutional neural network (CNN) approach to detect the soccer ball in an image sequence. In contrast to the existing methods where only the current frame or an image is used for the detection, we make use of the history of frames. Using history allows to efficiently track the ball in situations where the ball disappears or gets partially occluded in some of the frames. Our approach exploits spatio-temporal correlation and detects the ball based on the trajectory of its movements. We present our results with three convolutional methods, namely temporal convolutional networks (TCN), ConvLSTM, and ConvGRU. We first solve the detection task for an image using fully convolutional encoder-decoder architecture, and later, we use it as an input to our temporal models and jointly learn the detection task in sequences of images. We evaluate all our experiments on a novel dataset prepared as a part of this work. Furthermore, we present empirical results to support the effectiveness of using the history of the ball in challenging scenarios.
\end{abstract}


\section{Introduction}
\label{sec:intro}
The RoboCup introduced by Kitano et al. \cite{kitano1997robocup} serves as the central problem in understanding and development of Artificial Intelligence. The challenge aims at developing a team of autonomous robots capable of playing soccer in a dynamic environment. It requires the development of collective intelligence and an ability to interact with surroundings for effective control and decision making. Over the years several humanoid robots \cite{Farazi2013a,schnekenburger2017detection,ficht2018nimbro} have participated in the challenge.

One of the main hurdle identified within the tournament is perceiving the soccer ball. The efficient detection of soccer ball relies on how good the vision system performs in tracking the ball. For instance, consider cases where the ball disappears or gets occluded from robots point of view for a few frames. In such situations using the current frame is not useful. However, a dependence on the history of frames can help in making a proper move. In this work, we propose an approach which can effectively utilize the history of ball movement and improve the task of ball detection. We first utilize the encoder-decoder architecture of SweatyNet model and train it for detection of the ball in single images. Later we use it as a part of our proposed layers and learn from temporal sequences of images, thereby developing a more robust detection system. In our approach we make use of three spatio-temporal models: TCN~\cite{bai2018empirical}, ConvLSTM~\cite{xingjian2015convolutional} and ConvGRU~\cite{ballas2015delving}.

For this work, we recorded a new dataset for the soccer ball detection task. We make our data as well as our implementation available on GitHub so that the results can be easily reproduced. research~\footnote{\url{https://github.com/AIS-Bonn/TemporalBallDetection}}. We used Pytorch~\cite{paszke2017automatic} for our implementation.

\begin{figure}[h]
    \centering
    \includegraphics[width=0.65\textwidth, height=8.5cm]{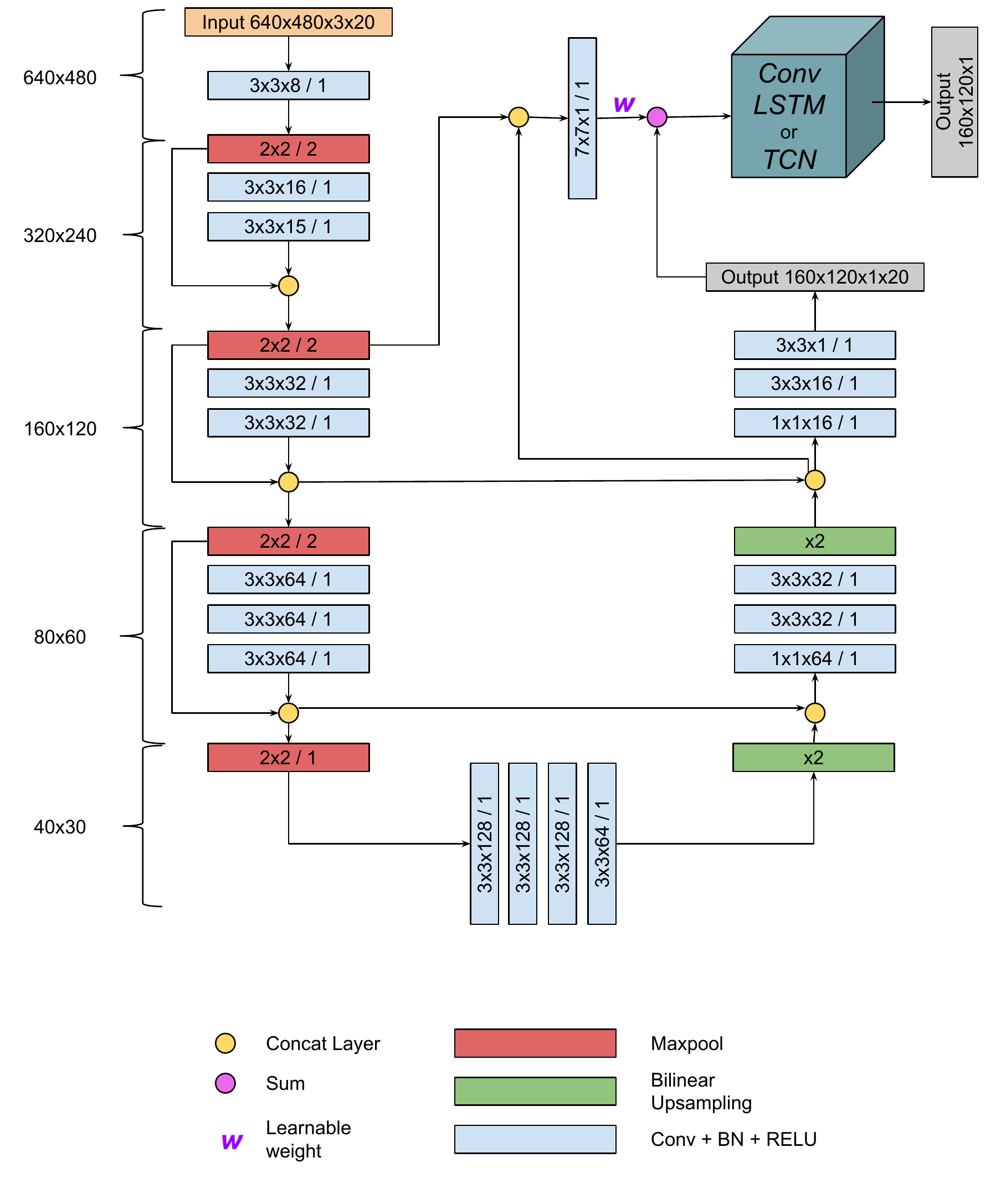}
    \caption{The proposed architecture with feed-forward and temporal parts.}
    \label{fig:architecture}
\end{figure}

\section{Related Work}
\label{sec:related}
Numerous works have been done in the area of soccer ball detection. Before RoboCup 2015 the ball was orange, and many teams used color information \cite{schulz2007ball}. Since RoboCup2015, the ball is not color coded anymore, which forced teams to use more sophisticated learning based approaches like HOG cascade classifier \cite{Farazi2015}. In recent years, the convolutional approaches with their innate ability to capture equivariance and hierarchical features in images have emerged as a favorite choice for the task.
In~\cite{speck2016ball} authors use CNN to perform localization of soccer ball by predicting the $x$ and $y$ coordinates. In a recent work~\cite{leiva2018playing} use proposal generators to estimate regions of soccer ball and further use CNN for the classification of regions.
In~\cite{javadi2017humanoid} authors compared various CNN architectures namely LeNet, SqueezeNet, and GoogleLeNet for the task of a ball detection by humanoid robots. In~\cite{schnekenburger2017detection} authors inspired by work of~\cite{ronneberger2015u} proposed a Fully Convolutional Networks (FCN) that offers a robust and low inference time, which is an essential requirement for the soccer challenge. As the name suggests, the FCN is composed entirely of convolution layers which allows them to learn a path from pixels in the first layers to the pixel in the deeper layers and produce an output in the spatial domain — hence making FCN architecture a natural choice for pixel-wise problems like object localization or image segmentation. In~\cite{houliston2018visual} authors use geometric properties of the scene to create graph-structured FCN. In~\cite{van2018deep} authors proposed a modification of U-Net~\cite{ronneberger2015u} architecture by removing skip connections from encoder to decoder and using depthwise separable convolution. This allows to achieve improvement in inference time and making it the right choice for real-time systems.

The existing work uses the current frame for the detection of the soccer ball. We hypothesize that the history of frames (coherent sequence of previous frames) could help model in making a better prediction, especially in cases where ball disappears or is missed for a few frames. To support our hypothesis, we extend our experiments and use temporal sequences of images. 
A crucial element of processing continuous temporal sequences is to encode consensual information in spatial and temporal domains simultaneously. There are several methods which allow extracting spatiotemporal video features like widely used Dense Trajectories~\cite{wang2013dense} where densely sampled points are tracked based on information from the optical flow field and describe local information along temporal and spatial axes.  In~\cite{carreira2017quo} authors proposed Two-Stream Inflated 3D ConvNet (I3D) where convolution filters expanded into 3D let the network learn seamless video feature in both domains. For predicting object movement in the video, Farazi et al. proposed a model based on frequency domain representation \cite{Farazi2019}. One of the recent methods in modeling temporal data is temporal convolution networks (TCN)~\cite{lea2017temporal}. The critical advantage of TCN is the representation gained by applying the hierarchy of dilated causal convolution layers on the temporal domain, which successfully capture long-range dependencies. Also, it provides a faster inference time compared to recurrent networks, which make it suitable for real-time applications.

Additionally, there are successful end-to-end recurrent networks which can leverage correlations within sequential data~\cite{xingjian2015convolutional,hochreiter1997long,chung2014empirical}. ConvLSTM~\cite{xingjian2015convolutional} and ConvGRU~\cite{ballas2015delving} are recurrent architectures which compound convolutions to determine the future state of the cell based on its local neighbors instead of the entire input. 

In this work, we propose a CNN architecture which utilizes sequences of ball movements in order to improve the task of soccer ball detection in challenging scenarios.


\section{Detection Models}
\label{sec:netarch}
\subsection{Single Image Detection}
In this paper, the task of soccer ball detection is formulated as a binary pixel-wise classification problem, where for a given image, the feed-forward model predicts the heatmap corresponding to the soccer ball. In this part we utilize three feed-forward models namely SweatyNet-1, SweatyNet-2 and SweatyNet-3 as proposed in~\cite{schnekenburger2017detection}.

All three networks are based on an encoder-decoder design. The SweatyNet-1 consists of five blocks in the encoder part and two blocks in the decoder part. In the encoder part, the first block includes one layer, and the number of filters is doubled after every block. In the decoder part, both blocks contain three layers. Each layer comprises of a convolutional operator followed with batch normalization and ReLU as the non-linearity. In addition, bilinear upsampling is used twice: after the last block of the encoder and after the first block of the decoder. Skip connections are added between layers of encoder and decoder to provide high-resolution details of the input to the decoder. Similar approaches have been successfully used in Seg-Net~\cite{badrinarayanan2017segnet}, V-Net~\cite{milletari2016v} and U-Net~\cite{ronneberger2015u}.

All convolutional filters across the layers are of the fixed size of $3\times3$. The encoder part includes four max-pooling layers where each one is situated after the first four blocks. The number of filters in the first layer is eight, and it is doubled after every max-pooling layer. In the decoder, the number of filters is reduced by a factor of two before every upsampling layer.  

The other two variants of SweatyNet, are designed to reduce the number of parameters and speed up the inference time. In SweatyNet-2, the number of parameters is reduced by removing the first layer in each of the last five blocks of the encoder. In SweatyNet-3, the number of channels is decreased by changing the size of convolutions to $1\times1$ in every first layer of last five encoder blocks and both of the decoder blocks.

\begin{figure}[h]
    \centering
    \begin{minipage}{.9\textwidth}
        \includegraphics[width=0.95\textwidth, height=0.24\textheight]{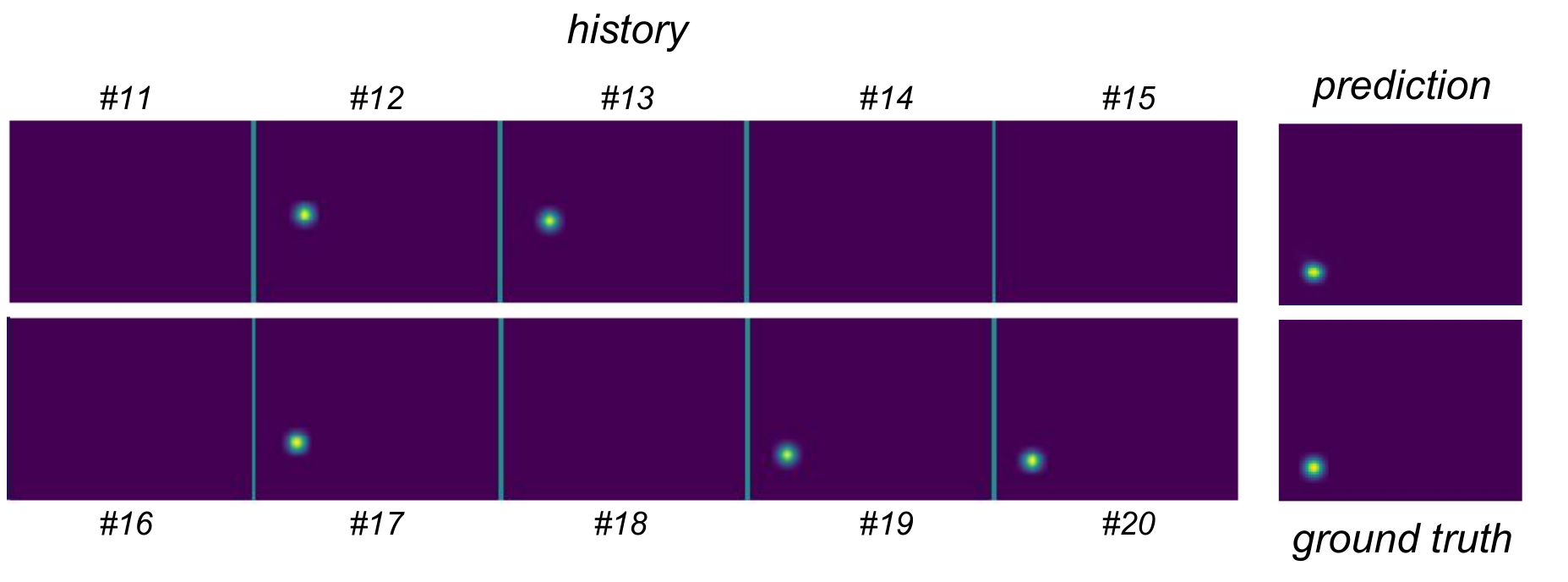}
    \end{minipage}
    \begin{minipage}{.07\textwidth}
        \includegraphics[width=0.8\textwidth, height=0.15\textheight]{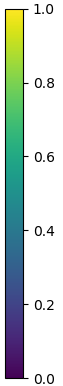}
    \end{minipage}
    \caption{The prediction results, on the synthetically generated sequences. The network correctly predicts the future position and successfully keeps the size of \textit{slow} moving ball with $\sigma=4$ even when the history is sparse. Note that sparse history resembles an occluded ball.}
    \label{fig:slow}
\end{figure}

\subsection{Detection in a Sequence}

\begin{figure}[h]
    \centering
    \includegraphics[width=0.98\textwidth]{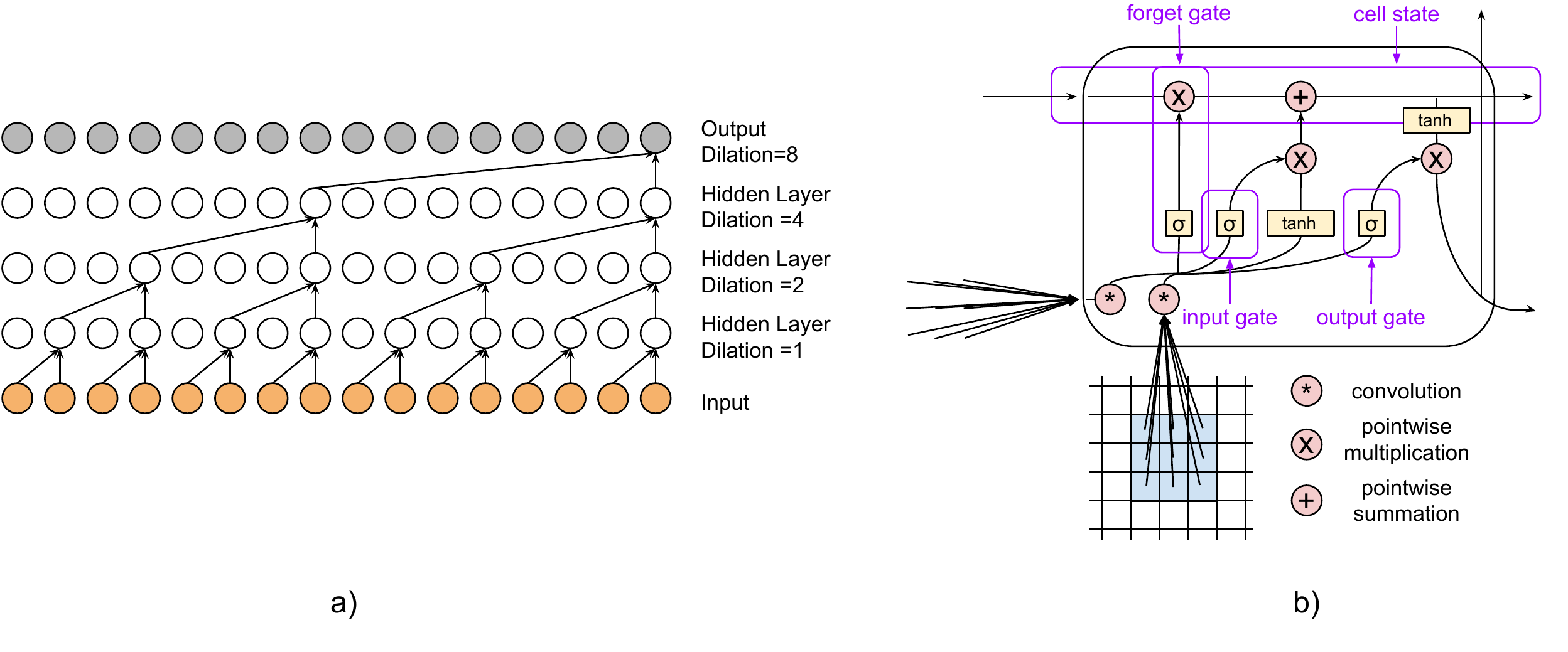}
    \caption{Visualization of a) a stack of causal convolutional layers which compose TCN architecture. b) a convolutional LSTM cell.}
    \label{fig:tcn_lstm}
\end{figure}

Temporal extensions capture spatio-temporal interdependence in the sequence and allow to predict the movement of the ball capturing its size, direction, and speed correctly. In our experiments, we utilize the temporal series of images to improve the task of soccer ball detection further. 

Our approach illustrated in Fig.~\ref{fig:architecture} propose a temporal layer and learnable weight $w$ which makes use of the history of sequences of fixed length to predict the probability map of the soccer ball. We use a feed-forward layer TCN and compare it with recurrent layers ConvLSTM and ConvGRU. The three approaches differ in the type of connections formed in the network.


We train our model to learn heatmaps of a ball based on the sequence of frames representing the history of its movement. More precisely, if the timestamp of the current frame is $t$, given the heatmaps from $(t-h)$ to $(t-1)$ the output of the network is the sequence of heatmaps from timestamp $t$ to $(t+p)$, where $h$ is the history length and $p$ is the length of predicted sequence.

The ConvLSTM and ConvGRU layers are stacks of several convolutional LSTM and GRU cells, respectively, which allows for capturing spatial as well as temporal correlations. Each ConvLSTM cell acts based on the \textit{input}, \textit{forget} and \textit{output} gates, while the core information is stored in the \textit{memory} cell controlled by the aforementioned gates. Each ConvGRU cell adaptively captures time dependencies with various time ranges based on \textit{content} and \textit{reset} gates.  Convolutional structure avoids the use of redundant, non-local spatial data and results in lower computations. Fig.~\ref{fig:tcn_lstm} depicts the structure of convolutional LSTM cell where the input is a set of flattened 1D array image features obtained with the convolutions layers. Convolutional GRU cell also differs from standard GRU cell only in the way how input is passed to it.

Unlike the two recurrent models, where gated units control hidden states, TCN hidden states are intrinsically temporal. This is attributed to the dilated causal convolutions used in TCN, which generates temporally structured states without explicitly modeling connection between them. Thus, TCN captures long term temporal dependencies in a simple feed-forward network architecture. This feature further provides an advantage of the faster inference time. Fig.~\ref{fig:tcn_lstm} shows dilated causal convolutions for dilations 1, 2, 4, and 8.
For our work, we replicated the original TCN-ED structure with repeated blocks of dilated convolution layers and normalized ReLU as activation functions.




For sequential data, it is challenging to train a network from scratch because of the limited size of the dataset and the difficulties in collecting the real data. Besides, the training process requires more memory to store a batch of sequences, resulting in a choice of smaller batch size. To address this problem, we use transfer learning and finetune the weights of our model on the sequences of synthetic data. We use SweatyNet-1 as the feature extractor and finetune it with the temporal layers. 

For the input to temporal layers; TCN, ConvLSTM, and ConvGRU, we also take advantage of high-resolution spatial information by concatenating the output of $2^{nd}$ and $6^{th}$ block of SweatyNet-1. To speed up the training process and propagate spatial information, we apply a convolution of size $7\times7$ on the combined features. Moreover, we take an element-wise product of the output of convolution with a learnable weight of $w$ and add it to the output of SweatyNet. This combination serves as an input to the temporal layers. The weight $w$ serves as a gate which learns to control how much of high-resolution information is transferred from the early layers of Sweaty-Net and helps the network in detecting soccer ball with subpixel level accuracy. 

\begin{figure}[h]
    \centering
    \begin{minipage}{.9\textwidth}
        \includegraphics[width=0.95\textwidth,height=0.24\textheight]{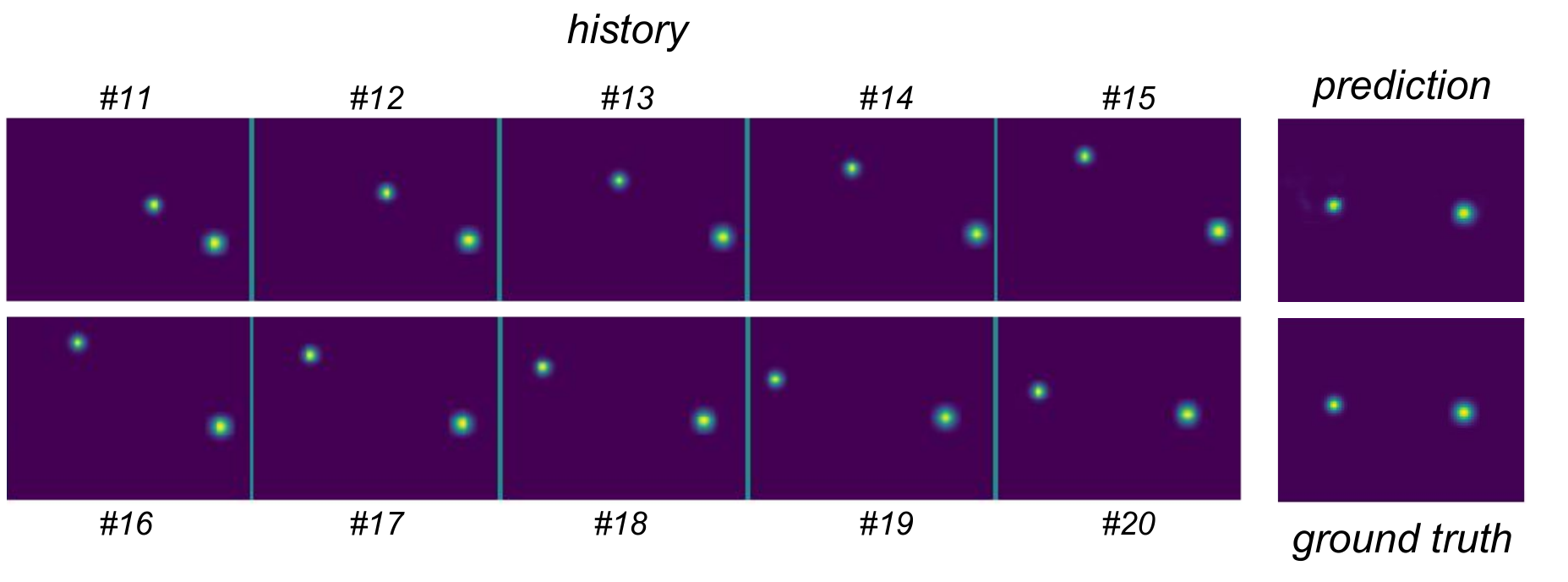}
    \end{minipage}
    \begin{minipage}{.07\textwidth}
        \includegraphics[width=0.8\textwidth, height=0.15\textheight]{pic/temp/colorbar.png}
    \end{minipage}
    \caption{The result of the temporal part, trained on a dataset with one ball per frame. Note that the network can generalize to detect two moving objects.}
    \label{fig:two}
\end{figure}

\begin{figure}[h]
    \centering
    \begin{minipage}{.9\textwidth}
        \includegraphics[width=0.95\textwidth,height=0.12\textheight]{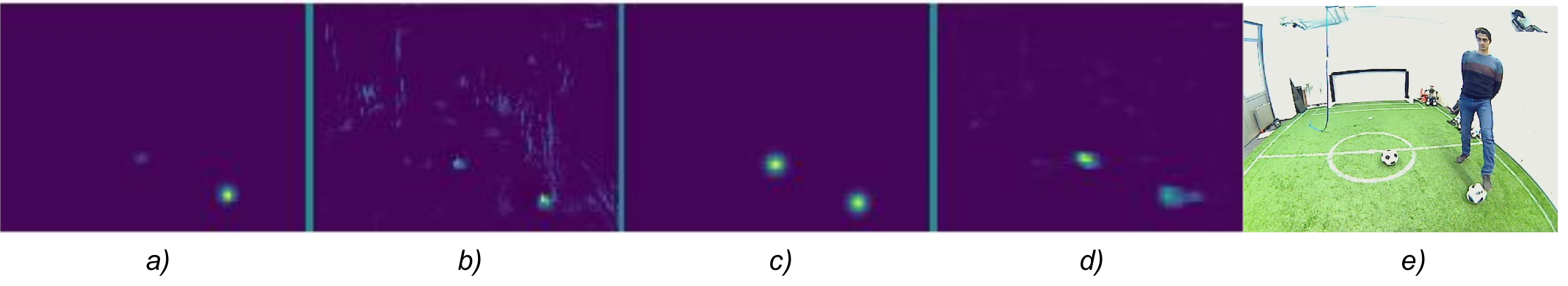}
    \end{minipage}
    \begin{minipage}{.07\textwidth}
            \includegraphics[width=0.8\textwidth,height=0.11\textheight]{pic/temp/colorbar.png}
    \end{minipage}
    \caption{Qualitative results of the trained network in detecting two balls. a) SweatyNet prediction b) residual information c) ground truth d) temporal prediction e) real image}
    \label{fig:two_real}
\end{figure}

\section{Experiments}
\label{sec:exp}
In this section, we describe the details of the training process for our two sets of experiments. In the first experiment, we consider a problem of localization of the object in an image. In the second experimet, we evaluated our temporal approach. The evaluation of our experiments is discussed in Section~\ref{sec:metric}. 

\subsection{Training}
\label{sec:train}
\noindent \textbf{Detection in an Image:} For our work, we created a dataset of 4562 images, of which 4152 images contain a soccer ball. We refer to it as \textit{SoccerData}. The images are extracted from a video recorded from the robot's point of view and are manually annotated using the imagetagger~\footnote{\url{https://imagetagger.bit-bots.de/}} library. The images are from three different fields with different light sources. Note that since the data is recorded on walking robot, in many images we have blurry data.

Each image is represented by a bounding box with coordinates: $x_{min},y_{min}, \\x_{max}, y_{max}$. For teaching signal we generated a binormal probability distribution centered at $c = 0.5 (x_{max}+x_{min}, y_{max}+y_{min})$ and with the variance of $r= 0.5 min(x_{max}-x_{min}, y_{max}-y_{min})$. In contrast to the work of (\cite{schnekenburger2017detection}) where authors consider ball of fixed radius, we take into account the variable radius of a ball by computing the radius based on the size of the bounding box.

We apply three variants of SweatyNet model as described in Section~\ref{sec:netarch} on the \textit{SoccerData}. For the fair evaluation of the model, we randomly split our data into $70\%$ training and $30\%$testing. In the training phase, mean squared error (MSE) is optimized between a predicted and a target probability map. We use Adam~\cite{kingma2014adam} as the optimizer. We trained all of our models for a maximum of 100 epochs on the Nvidia GeForce GTX TITAN GK110. Similar to (\cite{schnekenburger2017detection}) the hyperparameters used in our experiments are learning rate of $0.001$ and a batch size of $4$. In addition, we experiment with dropout probability of $0.0, 0.3$ and $0.5$.

\noindent \textbf{Detection in a Sequence:} 
We train the temporal part in two ways: (i) we pre-train the temporal model on artificially generated sequences and finetune it on top of the pre-trained SweatyNet-1 for the real sequences, \\
(ii) finetune the joint model on the real sequences where the pre-trained weights are used only for the SweatyNet-1 model.

Algorithm~\ref{alg:seq} details the procedure for synthetic data generation. To get heatmaps of a particular sequence $L_i$ at each time step $j$ we generate a multinormal probability distribution centered at $(x_j,y_j)$ with a variance equal to the radius $R_i$.

To finetune the model on the real sequential data, we extracted a set of real soccer ball frames from bags recorded during RoboCup2018 Adult-Size games. Since video frames do not always contain a ball in the field of view, we preprocess videos to make sure that we do not use a sequence of frames without any ball present. With such restrictions, we got 20 sets of consecutive balls with an average length of 60. For all of our experiments, we fixed the history size $h$ to 20 and prediction length $p$ to 1.

For training on real data, we use learning rates of $1e-5$ for the detection task and $1e-4$ for the temporal part after pretraining. In the temporal network on the artificial sequences, the learning rate is set to $1e-5$.  We train on synthetic data for 20 epochs and 30 epochs for the real data.

\noindent \textbf{TCN:} Encoder and decoder of TCN are two convolutional networks with two layers of 64 and 96 channels, respectively. We set up all parameters following the work of (\cite{lea2017temporal}) except that we use Adam as an optimizer with MSE loss. 

\noindent \textbf{ConvLSTM and ConvGRU:} We use four layers of ConvLSTM / ConvGRU cells with the respective size of 32, 64, 32, 1, and fixed kernel of size five across all layers.


\noindent \textbf{Multiple Balls in a Sequence:} To verify that our model can generalize, we test it on a more complex scenario with two present balls. Note that the network was only trained on a dataset containing a single ball.
The qualitative results can found in Fig.~\ref{fig:two} and Fig.~\ref{fig:two_real}. These figures depict that the model is powerful enough to handle cases not covered by training data. The temporal part leverages the previous frames and residual information and can detect the ball which is absent in SweatyNet output (Fig.~\ref{fig:two} a) vs. d)).

\subsection{Postprocessing}
\label{sec:post}
The output of a network is a probability map of size $160\times120$. We use the contour detection technique explained in Algorithm~\ref{alg:postproc} to find the center coordinates of a ball. The output of the network is of lower resolution and has less spatial information than the input image. To account for this effect, we calculate sub-pixel level coordinates and return the center of contour mass, as the center of the detected soccer ball.

\begin{algorithm}[h]
\caption{Postprocessing to find coordinates of a ball.}
\label{alg:postproc}
\begin{algorithmic}
\STATE $M\gets$ A matrix of size $120\times160$ representing predicted output of a network.\\
\STATE $A_{min}\gets$ Area of a smallest ball in the training set.
\STATE $B_M \gets$ $M>0.1$, binary map.\\
\STATE  List $L\gets$ detect contours in $B_M$.\\
\STATE  $C \gets$ contour in $L$ with maximum area.\\
\IF{area of $C\le A_{min}$.}\STATE $(c_x,c_y) \gets (-1,-1)$ 
\ELSE
\STATE  $W_C \gets$ mask $M$ with $C$.\\
\STATE  $(c_x,c_y) \gets$ center of mass of $W_C$.\\
\ENDIF
\RETURN $(c_x,c_y)$
\end{algorithmic}
\end{algorithm}

\begin{algorithm}[h]
\caption{Artificial sequences creation.}
\label{alg:seq}
\begin{algorithmic}
\STATE $N\gets$ Number of sequences.\\
\STATE $L_N \gets []$ A list to store a list of coordinates of a moving ball for all sequences.
\STATE $R_N \gets []$ A list to store radius of a ball for all sequences.
\FOR{i = 1 \textbf{to} $N$}
\STATE $R_i \sim \mathcal{U}(\{3,4,5\})$, $x_1 \sim \mathcal{U}(\{0,...,frame_x\}), y_1 \sim \mathcal{U}(\{0,...,frame_y\})$
\STATE $steps_i \sim \mathcal{U}(\{30,..., 60\})$ and $dist_i \sim \mathcal{U}(\{50,..., 500\})$
\STATE $dx_i\gets x_1 / steps_i, dy_i \gets y_1 / steps_i$ 
\STATE $R_N\gets R_i$, $L_i \gets [(x_1, y_1)]$ 
\FOR{j = 1 \textbf{to} $steps_i$}
\STATE $direction\_x_j, direction\_y_j \sim \mathcal{U}(\{-1, 1\})$ 
\STATE $x_j\gets x_j + direction\_x_j.dx_i$, $y_j\gets y_j + direction\_y_j.dy_i$
\STATE $L_i \gets (x_j, y_j)$
\ENDFOR
$L_N \gets L_i$
\ENDFOR
\RETURN $L_N, R_N$
\end{algorithmic}
\end{algorithm}

\begin{figure}[h]
    \centering
    \begin{minipage}{.9\textwidth}
        \includegraphics[width=0.9\textwidth]{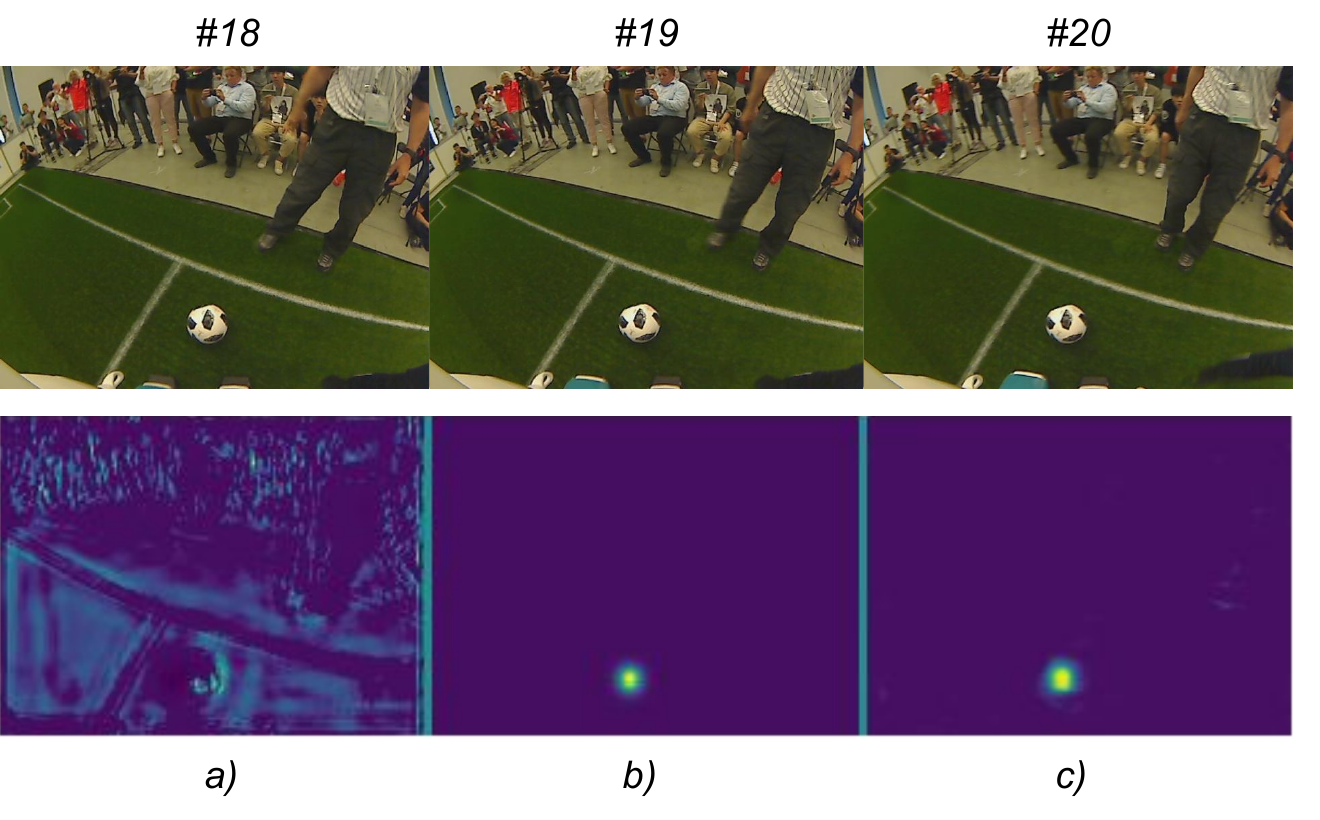}
    \end{minipage}
    \begin{minipage}{.07\textwidth}
        \includegraphics[width=0.8\textwidth, height=0.25\textheight]{pic/temp/colorbar.png}
    \end{minipage}
    \caption{Top row is the part of the input history (frame \{18,19,20\}). The bottom row consists of heatmaps where a) visualization of the residual information from Sweaty Net to temporal, b) ground truth ball position and c) predicted output by the temporal part.}
    \label{fig:temp}
\end{figure}

\begin{figure}[h]
    \centering
    \begin{minipage}{.9\textwidth}
        \includegraphics[width=0.9\textwidth]{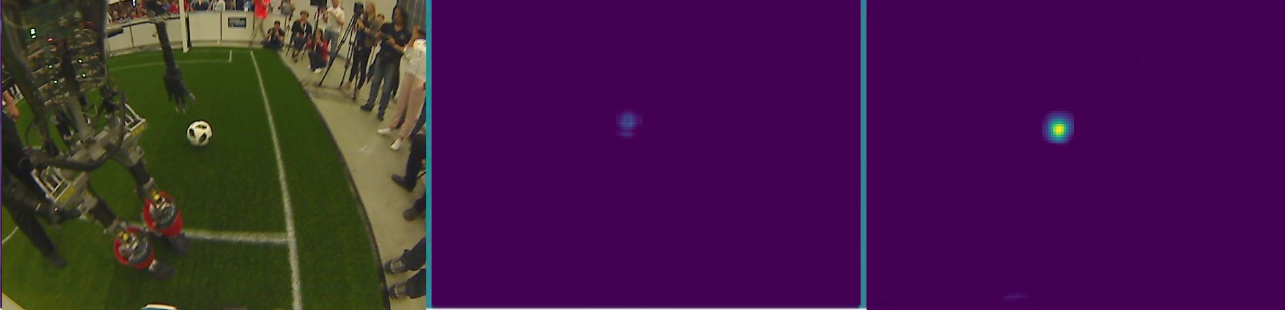}
    \end{minipage}
    \begin{minipage}{.07\textwidth}
            \includegraphics[width=0.9\textwidth, height=0.13\textheight]{pic/temp/colorbar.png}
    \end{minipage}
    \caption{Example of correctly detected ball after finetuning with the temporal model while the confidence of just the SweatyNet is very low, resulting in false negative detection. The left image is the real image; the middle is SweatyNet output without finetuning; the right one is SweatyNet output after finetuning.}
    \label{fig:temp_suc}
\end{figure}

\subsection{Evaluation}
\label{sec:metric}
To analyze the performance of different networks we use several metrics: false discovery rate (FDR), precision(PR), recall (RC), F1-score(F1) and accuracy (Acc) as defined in Eq.~\ref{eq:eq1}, where TP is true positives, FP is false positives, FN is false negatives, and TN is true negatives.
\begin{equation} \label{eq:eq1}
\begin{split}
&FDR = \frac{FP}{FP+TP}, PR = \frac{TP}{TP+FP}, RC = \frac{TP}{TP+FN},\\
&F1 = 2\times\frac{PR\times RC}{PR+RC}, Acc = \frac{TP+TN}{TP+FP+TN+FN} 
\end{split}
\end{equation}

An instance is classified as a TP if the predicted center and actual center of the soccer ball is within a fixed distance of $\gamma=5$.

\begin{figure}[h]
    \centering
    \begin{minipage}{.82\textwidth}
        \includegraphics[width=0.95\textwidth, height=7.0cm]{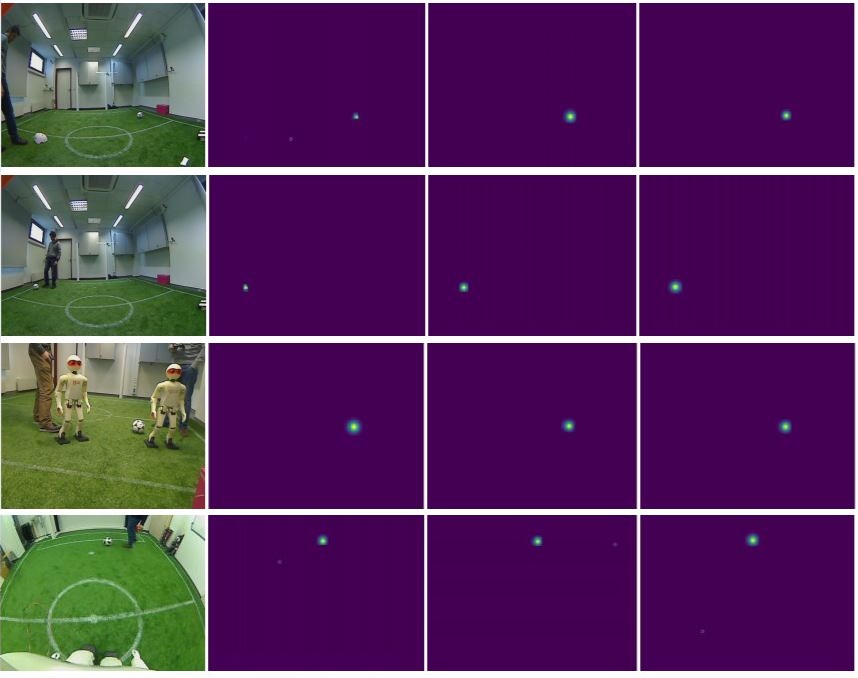}
    \end{minipage}
    \begin{minipage}{.07\textwidth}
        \includegraphics[width=0.8\textwidth, height=0.25\textheight]{pic/temp/colorbar.png}
    \end{minipage}
\caption{From left to right: input image, the ground truth, prediction of the neural network, and the final output after post-processing.}
    \label{fig:good}
\end{figure}


    \begin{table}
    \vspace{-4mm}
            \caption{Evaluation of SweatyNets on the task of soccer ball detection. The highlighted numbers are the best performance for a particular dropout probability. }
        \label{tab:table1}
        \centering
        \scriptsize
        \begin{tabular}{ccccccccccccc} 
            \toprule
            \multicolumn{12}{c}{\textbf{\textsf{Performance Metric}}}\\ 
            \cmidrule(r){1-12}
            \textbf{Model} & \textbf{Dropout} &\multicolumn{2}{c|}{\textbf{FDR}} & \multicolumn{2}{c|}{\textbf{PR}} & \multicolumn{2}{c|}{\textbf{RC}} & \multicolumn{2}{c|}{\textbf{F1}} & \multicolumn{2}{c}{\textbf{Accuracy}} \\ 
            \cmidrule(r){3-12}
            & &Train &Test| &Train &Test| &Train &Test| &Train &Test| &Train &Test \\ 
            \midrule
            SweatyNet-1& 0.0 &\textbf{0.001} &\textbf{0.017} & \textbf{0.999} & \textbf{0.981} &0.987 &\textbf{0.949} & 0.993 & \textbf{0.964} &0.989 &\textbf{0.945}\\
            SweatyNet-2 &0.0 &0.003 &0.020 & 0.997 & 0.980 &0.986 &0.912 & 0.992 & 0.948  &0.988 &0.916 \\
            SweatyNet-3 &0.0 &0.002 &0.019 & 0.998 & \textbf{0.981} & \textbf{0.990} &0.935 & 0.994 & 0.959 &\textbf{0.991} &0.933\\ 
            \midrule
            SweatyNet-1 &0.3 &0.017& \textbf{0.019} & 0.984 & 0.979 &\textbf{0.988} &0.950 & 0.986 & \textbf{0.966} &0.978 &\textbf{0.945} \\ 
            SweatyNet-2 &0.3 &\textbf{0.014} &0.022 & \textbf{0.986} & \textbf{0.980} &0.986 &0.949 & 0.986 & 0.964 &\textbf{0.979} &0.941\\
            SweatyNet-3 &0.3 &\textbf{0.014} &0.024 & \textbf{0.986} & 0.978 &0.987 &\textbf{0.956} & \textbf{0.987} & \textbf{0.966} &\textbf{0.979} &\textbf{0.945}\\ 
            \midrule
            SweatyNet-1 &0.5 &0.039 &0.024 & 0.960 & 0.975  &\textbf{0.989} &\textbf{0.972} & \textbf{0.974} & \textbf{0.973} &\textbf{0.961} &\textbf{0.955} \\
            SweatyNet-2 &0.5 &\textbf{0.029} &\textbf{0.015} & \textbf{0.970} & \textbf{0.983} &0.870 &0.812 & 0.917 & 0.899 &0.882 &0.844 \\
            SweatyNet-3 &0.5 &0.048 &0.022 & 0.952 & 0.981 &0.982 &0.940 & 0.967 & 0.959 &0.949 & 0.932\\ 
            \bottomrule
        \end{tabular}
        \vspace{-4mm}
    \end{table}

\section{Results}
The results of our experiments are summarized in Table~\ref{tab:table1}. The performance of all three models are comparable.
To improve generalization and prevent overfitting, we further experiment with different dropout~\cite{srivastava2014dropout} probability values. We train all our models on a PC with Intel Core i7-4790K CPU with 32 GB of memory and a graphics card Nvidia GeForce GTX TITAN with 6 GB of memory. For real-time detection, one major requirement is of a faster inference time. We report the inference time of the model on the NimbRo-OP2X robot in Table~\ref{tab:table2}(a). The NimbRo-OP2X robot is equipped with Intel Core i7-8700T CPU with 8 GB of memory and a graphics card Nvidia GeForce GTX 1050 Ti with 4 GB of memory. Since all three models don't use the full capacity of GPU during inference, which allows bigger models to perform extra computations in parallel; as a result, all three SweatyNet networks are comparable in real time inference. Fig.~\ref{fig:good} demonstrates the effectiveness of the model for the task of soccer ball detection. For this study, we only consider SweatyNet-1. 

The results of sequential part are further summarized in Table~\ref{tab:table2}(b). The sequential network successfully captures temporal dependencies and gives an improvement over the SweatyNet. Usage of artificial data for pre-training the temporal network is beneficial due to the shortage of real training data and boosts performance. Fig.~\ref{fig:slow} illustrates artificially generated ball sequences with the temporal prediction. We observed that when the temporal model is pre-trained on the artificial data, the learnable weight for the residual information takes a value of 0.57 on average, though without pre-training, the value is 0.49. The performance of TCN is comparable to ConvLSTM and ConvGRU, but it considerably outperforms ConvLSTM and ConvGRU in terms of inference time, which is a critical requirement for a real-time decision-making process. Table~\ref{tab:table2}(a) presents a comparison between temporal models on inference time.

To support our proposal of using sequential data, in Fig.~\ref{fig:temp_suc} we present an example image where the SweatyNet alone is uncertain of the prediction, though the network gives an strong detection when further processed with the temporal model.  
\begin{table}
\centering
        \caption{(a) Inference time comparison. For sequential models, we report time on top of the base model, (b) Evaluation of different tested models. Note \textit{real} denotes that training of the sequential part is performed only on real data and \textit{ft} means that a pre-training phase on the artificially generated ball sequences is done before finetuning on real data.}
        \label{tab:table2}
    \begin{tabular}{cccc}
    \toprule
        \textbf{Method} & \textbf{Time in ms measured on the robot}\\
        \midrule
        \midrule
         SweatyNet-1& 4.2\\
         SweatyNet-2& 3.5\\
         SweatyNet-3& 4.7\\ 
         \midrule
         LSTM & 219.6\\
         GRU & 178.5\\
         TCN & \textbf{1.1}\\
         \bottomrule
    \end{tabular}\\
\vspace{2mm}
 \text{(a)}\\
\vspace{4mm}
        \begin{tabular}{cccccc} 
            \toprule
            \multicolumn{6}{c}{\textbf{Performance Metric on Test set}}\\ \cmidrule{1-6}
            \textbf{Method} |& \textbf{FDR} |& \textbf{PR} |& \textbf{RC} |& \textbf{F1} |& \textbf{Acc} \\ 
            \midrule 
            SweatyNet-1 (0.5) & \textbf{0.024} & 0.975 & 0.972 & 0.973 & 0.955 \\
            Net+LSTM(\textit{real}) & 0.025 & \textbf{0.976} & 0.979 & 0.977 & 0.962 \\
            Net+LSTM(\textit{ft})& 0.026 & 0.975 & \textbf{0.987} & \textbf{0.981} & \textbf{0.967} \\
            Net+GRU(\textit{real}) & \textbf{0.024} & 0.975  & 0.980 & 0.978 & 0.963\\ 
            Net+GRU(\textit{ft})& 0.026 & 0.972  & \textbf{0.987} & \textbf{0.980} & \textbf{0.966}\\ 
            Net+TCN(\textit{real})& \textbf{0.024} & \textbf{0.976} & 0.982 & 0.979 & 0.964\\ 
            Net+TCN(\textit{ft})& 0.026 & 0.974 & \textbf{0.985} & \textbf{0.980} & \textbf{0.966} \\ 
            \bottomrule
        \end{tabular}\\
        \vspace{2mm}
        \text{b}\\
\end{table} 

\section{Conclusion}
In this paper, we address the problem of soccer ball detection using sequences of data. We proposed a model which utilizes the history of ball movements for efficient detection and tracking. Our approach makes use of temporal models which effectively leverage the spatio-temporal correlation of sequences of data and keeps track of the trajectory of the ball. We present three temporal models: TCN, ConvLSTM, and ConvGRU. The feed-forward nature of TCN allows faster inference time and makes it an ideal choice for real-time application of RoboCup soccer.
Furthermore, we show that with transfer learning, sequential models can further leverage knowledge learned from synthetic counterparts. Based on our results, we conclude that our proposed deep convolutional networks are effective in terms of performance as well as inference time and are a suitable choice for soccer ball detection. Note that the presented models can be used for detecting other soccer objects like goalposts and robots.

\bibliographystyle{splncs04}
\bibliography{main}

\end{document}